\documentclass[11pt,a4paper]{article}

\usepackage[utf8]{inputenc}
\usepackage[T1]{fontenc}
\usepackage{times}
\usepackage[margin=2.5cm]{geometry}
\usepackage{amsmath,amssymb,amsfonts}
\usepackage{mathtools}
\usepackage{graphicx}
\usepackage{booktabs}
\usepackage{array}
\usepackage{longtable}
\usepackage{xcolor}
\usepackage{caption}
\usepackage[round, authoryear]{natbib}
\setcitestyle{aysep={,}}
\usepackage[colorlinks=true,
            linkcolor=blue!70!black,
            citecolor=green!50!black,
            urlcolor=blue!70!black,
            breaklinks=true]{hyperref}

\usepackage{xspace}
\newcommand{\swave}{\textsc{SWave}\xspace}
\newcommand{\pam}{\textsc{PAM}\xspace}
\newcommand{\rwkv}{\textsc{RWKV}\xspace}
\newcommand{\mamba}{\textsc{Mamba}\xspace}

\newcommand{\survived}{\textcolor{green!50!black}{\textbf{Survived}}}
\newcommand{\reframed}{\textcolor{orange!80!black}{\textbf{Reframed}}}
\newcommand{\replaced}{\textcolor{blue!70!black}{\textbf{Superseded}}}
\newcommand{\decorative}{\textcolor{red!60!black}{\textbf{Not load-bearing}}}
\newcommand{\withdrawn}{\textcolor{red!70!black}{\textbf{Withdrawn}}}

\title{\texorpdfstring{%
  \textbf{Why \swave May Not Be All You Need:\\
  A Concept-Evolution Retrospective on\\
  Complex-Valued Recurrent Language Models}}{%
  Why SWave May Not Be All You Need:
  A Concept-Evolution Retrospective on
  Complex-Valued Recurrent Language Models}}

\author{
    Ramprasath Ganesaraja \\
    EdgeVerve Systems Limited \\
    \texttt{ramprasath.g@edgeverve.com}
    \and
    Swathika N \\
    EdgeVerve Systems Limited \\
    \texttt{swathika.n@edgeverve.com}
    \and
    Sahil Dilip Panse \\
    EdgeVerve Systems Limited \\
    \texttt{SahilDilip\_Panse@edgeverve.com}
}

\date{}

\begin{document}
\maketitle

% =============================================================================
\begin{abstract}
\swave is a complex-valued recurrent language model (169.26M parameters,
$D=384$, $L=16$, $T=2048$) trained on FineWeb-Edu using $2\times$H100~NVL.
It was designed around three founding premises: that representing
language as complex waves rather than real-valued numbers enables richer
information encoding; that a Cayley-parameterised unitary transition
provides a mathematical guarantee against state decay or explosion; and
that a hidden state which \emph{rotates} rather than shrinks preserves
signal integrity over arbitrarily long contexts.
The core of \swave evolved substantially across three development phases.
The Resonance Head was found to structurally admit imaginary-channel
collapse as a global loss minimum (a failure mode we term
\emph{cos-domination collapse}) and was superseded by an untied head
with independent real and imaginary embedding tables drawn from the
Phase-Associative Memory (\pam) architecture~\citep{vishwakarma2026pam}.
This resolved the degenerate minimum and enabled stable 200{,}000-step
training (best-step PPL~22.0 at step 89{,}861). ComplexNorm and the
Wave Propagation Scan proved load-bearing throughout all three phases and
were retained to the final architecture. ProtectGatedScan was reframed
as a structural prior rather than a learned behaviour. The four
multi-scale retention concepts, despite their design motivation, did
not produce differential cross-entropy improvement under controlled
evaluation and were found to be non-load-bearing. The ComplexGatedUnit
was superseded by a real-valued squared-ReLU channel mixer, which achieved equivalent performance
with fewer parameters. The auxiliary training objectives did not demonstrate measurable benefit
once the structural constraints they were designed to compensate were
resolved at the source.

The investigation yields a formal characterisation of cos-domination
collapse, a parallelisable scan with a log-space backward pass for
numerical stability, six transferable engineering principles for
complex-valued recurrent training, and a plan-to-code traceability
methodology for catching structural divergences that conventional test
suites miss. The documented concept lifecycle (what was retained, what
was reframed, what was superseded, and what proved non-load-bearing)
provides a reference case for future complex-valued model design.

\end{abstract}

% =============================================================================
\section{Introduction}
\label{sec:intro}

Transformer-based language models face two well-known scalability
constraints. First, the attention mechanism incurs $O(N^2)$ computation
and an $O(N)$-memory KV cache that grows linearly with context length,
making very long sequences economically prohibitive.
Second, the linear-recurrence alternatives (RWKV, Mamba/S4) that address
the quadratic cost do so through exponential decay: the state contracts
at each step, so early-context information fades over long sequences.
\swave was designed to escape both constraints simultaneously, matching
RWKV's $O(N)$ training cost and $O(1)$ inference memory while eliminating
the decay that causes SSMs to forget.

Recurrent sequence models with complex-valued hidden states offer a
theoretically motivated route to $O(1)$-memory inference with
norm-preserving long-range state retention \citep{arjovsky2016urnn, wisdom2016}.
The associative-memory interpretation draws on complex Hopfield networks
\citep{noest1992complex} and Holographic Reduced Representations
\citep{plate1995hrr}.
The unit-magnitude constraint
$h_t = e^{i\varphi}h_{t-1} + x_t$ preserves $\|h\|$ exactly across
arbitrarily long sequences, a property that real-valued recurrences
can only approximate through careful regularisation. Building on this
foundation, \swave set out to explore what a full-featured
complex-valued language model would look like: one that could match
Transformer-scale training while retaining $O(1)$ memory per sequence
position.

\paragraph{Founding premises.}
The design was motivated by three core ideas.
\textbf{Wave-based processing}: tokens are embedded as complex numbers,
so the hidden state carries both amplitude and phase, enabling richer
encoding than real-valued representations.
\textbf{Cayley unitary memory}: the state transition is Cayley-parameterised
to enforce $|\alpha_t|=1$, providing a formal guarantee that state energy
neither decays nor explodes across long sequences.
\textbf{Bounded-decay context}: because the state \emph{rotates} with
bounded decay rather than unconstrained exponential shrinkage at each
step, early-context signals are attenuated far less than in standard
real-valued recurrences, where decay is unbounded.
Sixteen design concepts were developed to realise this vision;
the paper documents what each became.

\paragraph{What we tried.}
Sixteen design concepts were developed across six groups: output head
(Resonance Head, Wave Embedding), state dynamics (Wave Propagation Scan,
AmplitudeGate, Unitary Rotation, Cayley Transform, ProtectGatedScan),
channel mixing (ComplexGatedUnit, WaveMixer), normalisation (ComplexNorm),
multi-scale retention (Wavelet State Hierarchy, Phase Bus, Echo Memory,
Resonant Router), and diagnostic infrastructure (Orthogonal
Initialisation, Wave Diagnostics, Wave Rewind). Each concept is presented
with its hypothesis and outcome in Section~\ref{sec:arch}. Development
proceeded in three phases (Table~\ref{tab:timeline}).

\paragraph{Development phases.}
Phase~1 (Original Idea) established the design concepts and the tied
resonance head architecture. Phase~2 (PAM Baseline) resolved a structural
issue in the output head by adopting \pam\ primitives (Phase~2 is a near-direct architectural adoption of \pam\ rather than an independent invention; \swave's specific contribution at this phase is empirical validation at 169M parameters and a vector-state rather than matrix-state design), ran a 200k-step
training run confirming stability, and refined the scan, normalisation,
and gradient monitoring infrastructure. Phase~3 (Integration) brought
the Phase~2 core back into the concept-rich Phase~1 architecture, evaluated
each retention concept on a stable foundation, and progressively replaced
Phase~2 primitives with \mamba/\rwkv\ load-bearing equivalents alongside
a falsifier-driven development methodology.

\begin{table}[ht]
\small\centering
\setlength{\tabcolsep}{5pt}
\begin{tabular}{p{3.0cm}p{3.4cm}p{7.0cm}}
\toprule
\textbf{Phase} & \textbf{Description} & \textbf{Focus} \\
\midrule
1.\ Original Idea &
  Tied phase embedding, resonance head, full set of design concepts &
  Cos-domination collapse first observed;
  five sequential rescue attempts, all reproducing the
  $245\!\times$ cos/sin ratio. \\
\midrule
2.\ PAM Baseline &
  PAM adoption; untied complex head; 200k-step training run &
  Untied head resolves degenerate real-only minimum;
  scan parallelisation; gradient and amplitude diagnostics;
  per-layer protect-bias schedule. \\
\midrule
3.\ Integration &
  Phase~2 core integrated into original Phase~1 architecture;
  Phase~2 primitives replaced by \mamba/\rwkv\ equivalents;
  falsifier methodology &
  Integration exposed which primitives were load-bearing.
  TwoStreamScan, ChannelMixBlock, RMSNorm, and real carrier
  adopted. Retention concepts evaluated on stable baseline. \\
\bottomrule
\end{tabular}
\caption{Three development phases of \swave.}
\label{tab:timeline}
\end{table}

\begin{figure}[ht]
\centering
\includegraphics[width=\linewidth]{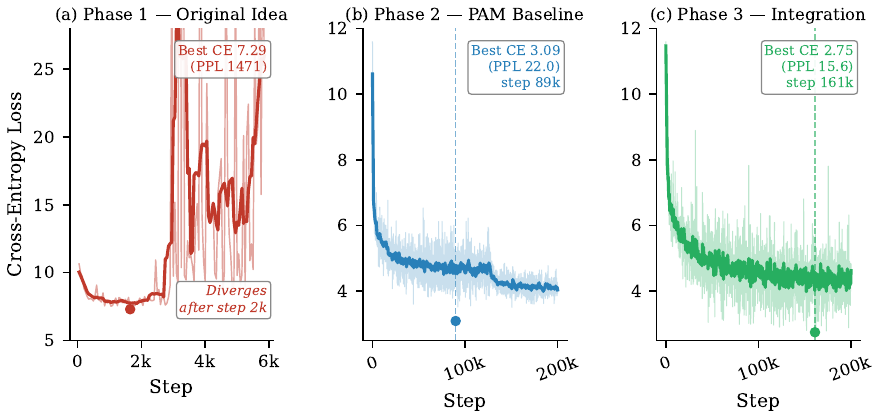}
\caption{Cross-entropy loss during training across the three development
phases.
\textbf{(a) Phase~1 (Original Idea):} The tied resonance head produces
unstable training that never exceeds CE~7.29 (PPL~1471) before diverging
to CE~25.3 by step~5{,}850, a signature of cos-domination collapse.
\textbf{(b) Phase~2 (PAM Baseline):} The untied head resolves the
degenerate minimum; training runs stably for 200{,}000 steps, reaching
best CE~3.09 (PPL~22.0).
\textbf{(c) Phase~3 (Integration):} The stable Phase~2 core integrated
into the original architecture reaches best CE~2.75 (PPL~15.6),
with noisier dynamics reflecting the more complex configuration.
Shaded curves are raw logs; solid curves are smoothed.}
\label{fig:training-curves}
\end{figure}

All results use a single model configuration: $D = 384$, $L = 16$,
$T = 2048$, $V = 100{,}277$ (cl100k\_base tokeniser), 169.26M parameters,
trained on FineWeb-Edu \citep{penedo2024fineweb} on $2\times$NVIDIA H100~NVL.

\paragraph{Verdict taxonomy.}
Each design concept is assigned one of five verdicts based on pre-specified
quantitative criteria (Table~\ref{tab:verdict-criteria}).

\begin{table}[ht]
\small\centering
\setlength{\tabcolsep}{4pt}
\begin{tabular}{p{2.4cm} p{10.9cm}}
\toprule
\textbf{Verdict} & \textbf{Criterion} \\
\midrule
\survived & On the final architecture default path; removal triggers a
  registered falsifier or a measurable PPL regression. \\
\reframed & Retained but original mechanistic claim updated based on
  empirical evidence; the concept survives in a revised form. \\
\replaced & Superseded by a structurally simpler or more principled
  alternative that achieves equivalent or better performance. \\
\decorative & Role subsumed by a core component, or requires conditions
  (e.g.\ frequency-structured priors, load-balancing signal) not present
  in the tested regime; the architectural investigation clarified this
  prerequisite. \\
\withdrawn & The specific capability was not measured in any documented
  run; the conditions required to evaluate it were not reached. \\
\bottomrule
\end{tabular}
\caption{Verdict taxonomy applied to all design concepts.}
\label{tab:verdict-criteria}
\end{table}

\paragraph{Scope.}
This paper does not claim a new state-of-the-art architecture. The value
lies in the documented analysis of why specific design decisions evolved
as they did, and the transferable methodology for identifying structural
divergences before they compound across multiple training runs.

% =============================================================================
\section{Architecture}
\label{sec:arch}

\swave is a complex-valued recurrent language model. Each hidden state in
Phase~2 is a complex vector $z \in \mathbb{C}^D$, stored as a pair of real
tensors $(z^r, z^i) \in \mathbb{R}^D \times \mathbb{R}^D$. The Phase~2
forward pass at each layer applies a sequence mixer (\texttt{ProtectGatedScan})
followed by a channel mixer (\texttt{ComplexGatedUnit}), with
\texttt{ComplexNorm} in a sandwich arrangement around each module.
Phase~3 integrated these Phase~2 components into the original Phase~1
architecture and subsequently replaced all four core primitives;
Section~\ref{subsec:phase3-training} documents those changes.

Each design concept below is presented with its original hypothesis,
followed by a unified account of how it evolved across the three phases,
and a final verdict.

% =============================================================================
\subsection{Output Head}
\label{subsec:output-head}

The output head maps the complex hidden state to a vocabulary distribution.

% ── Resonance Head ────────────────────────────────────────────────────────────
\subsubsection{Resonance Head (Tied Phase Head)}
\label{subsubsec:dc11}

\paragraph{Hypothesis.}
Each vocabulary item $v$ is assigned a learnable phase $\theta_v \in \mathbb{R}$;
the logit is:
\begin{equation}
\ell_v(h_r, h_i;\,\theta_v) = \cos(\theta_v)\,h_r + \sin(\theta_v)\,h_i
= \cos(\theta_h - \theta_v),
\label{eq:resonance-head}
\end{equation}
where the final equality holds when $h = e^{i\theta_h}$ is unit-norm.
Vocabulary retrieval was conceived as a phase-alignment operation: tokens
whose learned phase is close to the hidden state's current phase score
high, analogous to resonance in a tuned oscillator.

\paragraph{Journey.}
The first training run (Phase~1) reproduced a predicted failure mode at
step~2{,}000: the cos/sin term ratio at the output head reached $245\times$,
vocabulary phases were essentially frozen ($\theta_\mathrm{drift,mean} = 6.7
\times 10^{-4}$ rad), and cross-entropy showed no monotone descent over any
500-step window. Formal analysis established why this is unavoidable under
the tied parameterisation: for any token distribution expressible by $h_r$
alone, the configuration $(h_r,\,h_i=0)$ attains the same cross-entropy
minimum as any $h_i \neq 0$ configuration, because the gradient
$\partial\mathcal{L}_\mathrm{CE}/\partial h_i = \sum_v \sin(\theta_v)(p_v - y_v)$
vanishes as $\sin(\theta_v) \to 0$ for dominant vocabulary items.
The tied constraint $\cos^2\theta_v + \sin^2\theta_v = 1$ makes the
$h_i \equiv 0$ sub-manifold a global, not local, minimum.
Five successive interventions targeting optimisation dynamics confirmed
that the failure was architectural rather than optimisation-dependent:
gradient clipping, warmup adjustment, learning rate reduction, auxiliary
losses, and initialisation changes each reproduced the $245\times$
signature without altering the loss landscape geometry.

The structural resolution came from \pam\ \citep{vishwakarma2026pam}
in Phase~2: replacing the tied head with two independent real matrices
$E_r, E_i \in \mathbb{R}^{V \times D}$, initialised with $\mathcal{N}(0, 0.02^2)$,
so that the logit becomes $\ell_v = E_r[v]^\top h^r + E_i[v]^\top h^i$.
This drops the head-term initialisation ratio from $775\times$ to
${\approx}1.0\times$, giving both channels equal logit variance from step~0.
The 200k-step training run confirmed structural stability: $\rho =
\mathrm{RMS}(h_i)/\mathrm{RMS}(h_r) \in [0.79, 1.22]$ throughout.
In Phase~3, a \texttt{PhaseAttentionHead} variant with three learned
bridge projections $(W_\rho, W_{\varphi^r}, W_{\varphi^i})$ was
developed and verified via plan-to-code audit; the plain untied head
was confirmed as the production default.

\begin{figure}[ht]
\centering
\includegraphics[width=\linewidth]{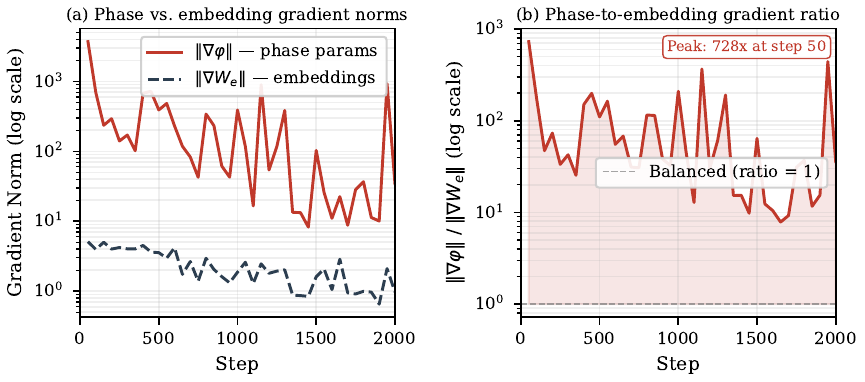}
\caption{Cos-domination collapse signature in Phase~1 training logs.
\textbf{(a)} The phase-parameter gradient norm
$\|\nabla\varphi\|$ (red) exceeds the embedding gradient norm
$\|\nabla W_e\|$ (dark) by orders of magnitude throughout the early
steps, indicating the loss surface is almost entirely shaped by the
phase parameters.
\textbf{(b)} The phase-to-embedding gradient ratio peaks at $728\times$
at step~50 and remains chronically elevated, confirming that the tied
head's loss landscape structurally drives $h_i \to 0$ regardless of
auxiliary rescues. The degenerate minimum is architectural, not
optimisation-level.}
\label{fig:cos-domination}
\end{figure}

\paragraph{Verdict.}
\replaced. The tied resonance head is superseded by the untied \pam\ head.
The cos-domination collapse analysis is the primary finding of this work,
empirically observed in training and formally characterised.

% ── Wave Embedding ────────────────────────────────────────────────────────────
\subsubsection{Wave Embedding (Tied Unit-Circle Embedding)}
\label{subsubsec:dc1}

\paragraph{Hypothesis.}
Each token $v$ is embedded as a point on the unit circle:
$h^r = \cos\theta_v$, $h^i = \sin\theta_v$, with
$\cos^2\theta_v + \sin^2\theta_v = 1$ enforced per (token, channel).
The two components carry distinct semantic roles: amplitude encodes
the importance or salience of a token, while phase encodes its
semantic direction, with semantically related tokens expected to
occupy nearby angles.
This wave-like representation was expected to support constructive
interference (related concepts reinforcing each other in the hidden
state) and destructive interference (noise and irrelevant signals
cancelling) through the phase relationships between complex-valued
representations.
Embeddings living on a phase manifold were also expected to align
naturally with the resonance head's phase-matching retrieval.

\paragraph{Journey.}
The unit-circle constraint was conceived in Phase~1 as the input-side
complement to the resonance head. The two concepts were tightly coupled:
the constraint meant embeddings had magnitudes of exactly 1, which
combined with the tied head's cos/sin parameterisation to make
$h_i \equiv 0$ a global rather than local loss minimum; the constraint
removed the escape route that free magnitudes would have provided.
When the resonance head was replaced by the untied head in Phase~2,
the unit-circle constraint was simultaneously released: the independent
tables $E_r, E_i \in \mathbb{R}^{V \times D}$ carry unconstrained
magnitudes, and the structural pathology dissolved. No change was made
in Phase~3.

\paragraph{Verdict.}
\withdrawn. The amplitude/phase semantic framing (amplitude encoding
token salience, phase encoding semantic direction) is a coherent
design hypothesis that was not independently evaluated. The unit-circle
magnitude constraint became structurally entangled with the tied
resonance head: releasing it was part of the fix for that head, not a
rejection of the embedding concept itself. Whether free-magnitude
complex embeddings with explicit phase priors provide richer
representations than standard real-valued embeddings remains an open
question.

% =============================================================================
\subsection{State Dynamics: The Recurrent Core}
\label{subsec:state}

The heart of \swave is a first-order linear recurrence over a complex hidden
state. Four design concepts shaped how that recurrence was gated, rotated,
and implemented efficiently.

% ── Wave Propagation Scan ─────────────────────────────────────────────────────
\subsubsection{Wave Propagation Scan}
\label{subsubsec:dc6}

\paragraph{Hypothesis.}
A complex first-order linear recurrence $h_t = A_t \odot h_{t-1} + x_t$,
with $A_t$ input-dependent and parallelisable via the associative operator
$(a_1, b_1) \oplus (a_2, b_2) = (a_1 a_2,\; a_2 b_1 + b_2)$.
$O(\log T)$ depth parallel prefix computation was expected to replace
sequential $O(T)$ depth, enabling efficient GPU training at full sequence
length.

\paragraph{Journey.}
The Phase~1 implementation used a sequential Python loop over $T=2048$
timesteps, dispatching ${\sim}150{,}000$ CUDA kernels per forward pass and
achieving only $40$--$42\%$ GPU utilisation due to dispatch overhead.
Phase~2 brought three successive improvements that together made the scan
practical at scale: first, a chunked scan with chunk size $C=64$ reduced
the loop from $T$ to ${\approx}96$ iterations; second, a closed-form
vectorisation of the intra-chunk recurrence using cumulative products,
\begin{align}
c[t] &= \prod_{s=1}^{t} r[s] \quad (\texttt{cumprod}), \\
h[t] &= c[t] \cdot \mathrm{cumsum}\!\left(\frac{x[t]}{c[t]}\right),
\end{align}
reduced kernel launches per scan call from ${\sim}158$ to $6$ (a $26\times$
reduction), recovering GPU utilisation from $40$--$42\%$ to near-peak;
third, the scan body is formulated in log-space so that all cumulative
products stay in $(0,1]$, bounding the largest backward intermediate by
$\exp(0)=1$ and avoiding fp32 overflow at the production sequence length
and halflife range. Log-$\alpha$ is clamped at $\log(1-10^{-5})$ to
prevent decay saturation at long halflives, verified by a
machine-verifiable falsifier (\texttt{atan2} gradient finite at origin).
Both requirements were identified during numerical analysis of the
fp32 dynamic range at $T=2048$.
The associative scan operator carried forward into Phase~3 unchanged,
generalised to the two-stream setting of \texttt{TwoStreamScan}
(Section~\ref{subsec:phase3-training}).

\paragraph{Verdict.}
\survived. The associative scan operator is load-bearing throughout all
three phases; the parallelisation improvements are transferable to any
first-order linear recurrence.

% ── AmplitudeGate ─────────────────────────────────────────────────────────────
\subsubsection{AmplitudeGate}
\label{subsubsec:dc3}

\paragraph{Hypothesis.}
A standalone magnitude gate controlling write strength:
$\mathrm{gate}_t = \sigma(W_g \cdot x_t)$, $h_\mathrm{gated} = \mathrm{gate}_t \odot h_t$,
with decay hardcoded at $1.0$. Selective write suppression was expected to
prevent uninformative inputs from polluting the hidden state.

\paragraph{Journey.}
In Phase~1, the AmplitudeGate operated as a standalone post-recurrence
module. Its core limitation became apparent: with decay fixed at $1.0$,
the gate could suppress writes but could not decelerate state decay,
so state preservation under uninformative inputs was incomplete.
Phase~2 addressed this with the \texttt{ProtectGatedScan}, which unified
the gating function into the recurrence itself via a dual-purpose protect
gate $p_t = \sigma(W_p|z_t| + b_p)$ with $b_p = -3.0$:
\begin{align}
\gamma_t &= e^{-\Delta t_t}(1-p_t) + p_t, \\
V'_t     &= V_t \cdot (1-p_t).
\end{align}
The same $p_t$ simultaneously suppresses the write ($V'_t$) and
decelerates decay ($\gamma_t$), so high $p_t$ freezes the state rather
than merely limiting the write, providing a strictly more complete form of state
preservation than the original gate. The $b_p = -3.0$ bias initialises
$p_t \approx 0.047$, biasing toward preservation from step~0.
The ProtectGatedScan was itself superseded by \texttt{TwoStreamScan}
in Phase~3, but the dual-purpose gating insight carried forward into
the new design.

\paragraph{Verdict.}
\replaced. AmplitudeGate's conceptual contribution, explicit write
suppression, was subsumed and generalised by the dual-purpose protect gate.

% ── Unitary Rotation ──────────────────────────────────────────────────────────
\subsubsection{Unitary Rotation}
\label{subsubsec:dc4}

\paragraph{Hypothesis.}
Step-wise unitary rotation of the hidden state:
\begin{align}
h^r_\mathrm{new} &= \cos\varphi\,h^r - \sin\varphi\,h^i + x^r, \\
h^i_\mathrm{new} &= \sin\varphi\,h^r + \cos\varphi\,h^i + x^i.
\end{align}
Exact unitary transitions were expected to prevent gradient
vanishing/explosion \citep{arjovsky2016urnn} and encode positional
information as a continuous rotation in the complex plane.

\paragraph{Journey.}
The Phase~1 implementation applied the rotation as a standalone module at
each recurrence step. The additive $+x$ term entangled the rotation with
the write operation, making the update difficult to interpret or analyse
independently. In Phase~2, the rotation was factored out of the recurrence
and applied as a phase hook on the key tensor $K_t$ before the conjugate
write; the formula $\cos(m\theta_d)z^r - \sin(m\theta_d)z^i$ is
mathematically identical but cleanly separated from the gating logic,
and has a clear positional-encoding interpretation analogous to
RoPE \citep{su2021rope}. This factored form carried into Phase~3 as a
structural prior in the log-$\alpha$ multi-timescale spectrum.

\paragraph{Verdict.}
\survived\ (reframed). The phase-preserving constraint survives in
\texttt{ComplexNorm} and in bounded $\gamma_t$; the original ``zero decay''
claim is reframed as ``phase-preserving with bounded decay,'' which is
what the implementation actually delivers.

% ── Cayley Transform ──────────────────────────────────────────────────────────
\subsubsection{Cayley Transform for Unitary Matrices}
\label{subsubsec:dc5}

\paragraph{Hypothesis.}
Parameterise unitary weight matrices via the Cayley map:
$U = (I - A)(I + A)^{-1}$, where $A$ is skew-symmetric.
The Cayley map covers the full unitary group without the rank deficiency
of matrix exponential approximations, offering exact parameterisation
of all unitary transformations.

\paragraph{Journey.}
When evaluated in Phase~1, the Cayley map required a matrix inversion
$(I+A)^{-1}$ at $D = 384$, costing $O(D^3)$ per forward pass
(${\sim}57M$ FLOPs at $D=384$; ${\sim}68B$ FLOPs at $D=4096$) and
exhibiting numerical sensitivity as $(I+A)$ approaches singular.
A diagonal approximation (restricting $A$ to a skew-diagonal, i.e.\
scalar rotation angles per channel) reduces this to $O(D)$
($384$ multiplications instead of $57M$), recovering the zero-decay
property at negligible cost; however, it also removes the inter-channel
mixing that full unitarity provides.
In Phase~2, the full-matrix path was replaced by a spectral
decomposition $R^m = V\,\mathrm{diag}(e^{i\omega m})\,V^H$, computed via
\texttt{torch.linalg.eigh} on the Hermitian generator
$H = -iK$, $K = (A - A^H)/2$, with exact unitarity guaranteed and fully
differentiable through PyTorch. However, across all phases the strict
unitary constraint proved less useful in practice than the bounded decay
with a learned protect gate, which achieves near-unitary behaviour where
appropriate without the computational overhead of a full matrix
decomposition.

\paragraph{Verdict.}
\decorative. The spectral form captures the stability benefit of the
Cayley formulation efficiently; the full-matrix unitary constraint
is not required to realise that benefit in the tested regime, pointing
to the spectral approximation as the operative principle.

% =============================================================================
\subsection{Channel Mixing: The Per-Position Nonlinearity}
\label{subsec:channel}

Between recurrent state updates, the model applies a per-position
nonlinear transformation. Two design concepts shaped this.

% ── Channel Mixer ─────────────────────────────────────────────────────────────
\subsubsection{\texorpdfstring{Channel Mixer (SwiGLU FFN $\to$ ComplexGatedUnit)}{Channel Mixer (SwiGLU FFN to ComplexGatedUnit)}}
\label{subsubsec:dc18}

\paragraph{Hypothesis.}
The standard gated FFN from \citet{shazeer2020glu}:
$(W_1 x \odot \mathrm{SiLU}(W_2 x)) \cdot W_3$.
Using a well-validated real-valued nonlinearity was expected to allow
borrowing published hyperparameter settings and provide a stable
channel mixer from the outset.

\paragraph{Journey.}
In Phase~1, applying SwiGLU to complex $z = (z^r, z^i)$ by
channel-splitting (treating real and imaginary parts as independent
real vectors) broke the complex multiplication structure and discarded
the phase relationships the rest of the architecture was designed to
preserve. Phase~2 replaced SwiGLU with the \texttt{ComplexGatedUnit}
(CGU), a five-step operation that works natively on complex inputs:
\begin{align}
z_\mathrm{up}  &= W_\mathrm{up}\,z, \\
z_\mathrm{act} &= \mathrm{modReLU}(z_\mathrm{up}), \\
z_\phi         &= W_\phi\,z\;/\;|W_\phi\,z|, \\
z_\mathrm{rot} &= z_\mathrm{act} \odot z_\phi, \\
\mathrm{CGU}(z) &= W_\mathrm{down}\!\left(z_\mathrm{rot} \odot \sigma(|W_g z|)\right),
\end{align}
where $\mathrm{modReLU}(z) = \mathrm{ReLU}(|z|-b)\cdot z/|z|$
preserves phase while thresholding magnitude, $z_\phi$ is a unit-phase
gate rotating the activated state in $\mathbb{C}$, and $\sigma(|W_g z|)$
is a real-valued magnitude gate. In Phase~3, as the hidden-state carrier
shifted from complex to real, CGU was in turn superseded by
\texttt{ChannelMixBlock} (RWKV-V5 channel mix \citep{peng2023rwkv}),
which is structurally matched to the real carrier.

\paragraph{Verdict.}
\replaced. Each transition (SwiGLU to CGU, CGU to ChannelMixBlock) reflects
a tightening alignment between the channel mixer and the carrier type.

% ── WaveMixer ─────────────────────────────────────────────────────────────────
\subsubsection{WaveMixer (Token-Shift Blend)}
\label{subsubsec:dc2}

\paragraph{Hypothesis.}
Blend current and previous tokens before the channel mixer:
$x_k = \mu_k \odot x_\mathrm{cur} + (1-\mu_k) \odot x_\mathrm{prev}$,
with $\mu_k, \mu_v, \mu_r$ learnable per-channel blend scalars and
a gated output $= \sigma(r) \times \tanh(k) \times v$, following the
RWKV-V4 token-shift approach \citep{peng2023rwkv}.
Learnable blend weights were expected to encode optimal per-channel
mixing as training progressed.

\paragraph{Journey.}
Designed in Phase~1 as a dynamic blend mechanism, the WaveMixer was
not active in the Phase~2 PAM-aligned configuration. When re-introduced
in Phase~3 with asymmetric initialisation $(\mu_k, \mu_v, \mu_r) = (0.3, 0.5, 0.7)$
to break the symmetry of the three-vector parametrisation, the
$\mu$ vectors remained bit-identical to initialisation across 500 steps
and all 16 blocks. The mechanism did not learn; the asymmetric
initialisation itself acted as a structural prior encoding a fixed blend
rather than a learned one. Phase~3 E7 promotion removed the token-shift
vectors from the default model graph.

\paragraph{Verdict.}
\reframed\ $\to$ \replaced.\footnote{The arrow notation indicates a concept that was initially reframed (the asymmetric initialisation was understood as a structural prior) and subsequently removed from the default path entirely, warranting the stronger Superseded verdict.} The dynamic blend hypothesis is not supported
by the training evidence; the asymmetric initialisation functions as a
static structural prior. Removed from the default path.

% =============================================================================
\subsection{Normalisation: Protecting Phase Geometry}
\label{subsec:norm}

% ── ComplexNorm ───────────────────────────────────────────────────────────────
\subsubsection{ComplexNorm}
\label{subsubsec:dc12}

\paragraph{Hypothesis.}
Standard RMSNorm applied independently to real and imaginary parts would
distort the phase relationship $\angle z = \mathrm{atan2}(z^i, z^r)$
by scaling the two components by different factors. \texttt{ComplexNorm}
normalises by the joint complex magnitude,
\begin{equation}
\mathrm{rms}(z) = \sqrt{\mathrm{mean}_{d}(|z_d|^2) + \varepsilon},
\qquad
\tilde{z}_d = s_d \cdot z_d\,/\,\mathrm{rms}(z),
\label{eq:complexnorm}
\end{equation}
with learnable per-channel gain $s \in \mathbb{R}^D$ (init $1.0$),
preserving $\angle\tilde{z}_d = \angle z_d$ by construction.

\paragraph{Journey.}
ComplexNorm was designed and implemented in Phase~1 as the normalisation
layer for the complex carrier. Its critical role became clearer in Phase~2,
when engineering investigations revealed that unconstrained amplitude
growth was driving phase-gradient instability: at step~37{,}250,
phase drift was negligible ($\theta_\mathrm{drift,mean} = 0.00144$ rad)
while the phase-group gradient norm escalated to $150$--$290$, because
phase gradients scale as $|h|^2$. Promoting ComplexNorm to a sandwich arrangement, applied before each
sublayer and after each residual add, bounded residual magnitudes at
every layer. Without this, geometric amplification of
${\sim}1.5\times$/layer compounds to $1.5^{16} \approx 656\times$
across a 16-layer stack. The sandwich-norm pattern is established in
Stable LM 3B and Gemma~2 \citep{gemmateam2024gemma}; \swave's
contribution is the quantitative diagnosis specific to the complex
carrier, where phase-gradient scaling as $|h|^2$ makes amplitude
control essential.
In Phase~3, when the hidden-state carrier shifted from complex to real,
\texttt{ComplexNorm} was replaced by standard real-valued \texttt{RMSNorm}
as the appropriate normalisation for the real carrier;
the sandwich arrangement was retained throughout.

\paragraph{Verdict.}
\survived. Phase-preserving normalisation is load-bearing for the
complex carrier; the sandwich arrangement is a transferable engineering
principle retained across all three phases.

% =============================================================================
\subsection{Multi-Scale Retention}
\label{subsec:retention}

Four design concepts were developed to give \swave richer multi-scale
memory, enabling the model to simultaneously reason over short,
medium, and long temporal contexts within a single recurrent pass.
All four were evaluated in the Phase~3 controlled ablation
and found to be non-load-bearing.

% ── Wavelet State Hierarchy ───────────────────────────────────────────────────
\subsubsection{Wavelet State Hierarchy}
\label{subsubsec:dc7}

\paragraph{Hypothesis.}
Maintain hidden states at three temporal strides $\{1, 4, 16\}$ and blend
them with an input-dependent softmax router:
$\mathrm{blend} = \mathrm{softmax}(W \cdot [|h_f|, |h_m|, |h_c|])$.
Motivated by wavelet decomposition theory, different stride levels were
expected to capture different temporal scales of the input signal,
mirroring the hierarchical temporal structure of natural language.

\paragraph{Journey.}
Designed in Phase~1 and retained as an opt-in flag in Phase~2, the
Wavelet State Hierarchy was evaluated in a controlled 500-step ablation
in Phase~3 alongside the other three retention concepts. The CE descent
curves showed no differential improvement across the evaluation window.
The investigation pointed to a missing inductive bias: multi-stride states
without frequency-structured priors (cf.\ HiPPO initialisation
\citep{gu2020hippo} in Mamba) do not constitute a multi-scale mechanism.
Different strides provide different receptive fields but no structural
frequency bias, leaving the router without a substantive signal to
route on.

\paragraph{Verdict.}
\decorative. Multi-stride states require frequency-structured initialisation
(e.g.\ HiPPO-style priors) to give each stride a distinct functional
role; the investigation identified this as the missing inductive bias.

% ── Phase Bus ─────────────────────────────────────────────────────────────────
\subsubsection{Phase Bus (Cross-Layer EMA Communication)}
\label{subsubsec:dc8}

\paragraph{Hypothesis.}
A cross-layer communication channel propagating a phase signal via
exponential moving average:
$\mathrm{bus} \leftarrow \mathrm{ema} \cdot \mathrm{bus} + (1-\mathrm{ema}) \cdot \mathrm{scale} \cdot h$,
$h \leftarrow h + \mathrm{read\_scale} \cdot \mathrm{bus}$,
with learnable per-block write and read scales.
Deeper layers injecting a summary phase signal into earlier layers was
expected to enable global coherence across the stack at $O(1)$ cost.

\paragraph{Journey.}
Designed in Phase~1 and retained as opt-in in Phase~2, the Phase Bus was
evaluated in Phase~3 and showed no differential improvement. The residual
stream already handles cross-layer communication effectively at this
scale. Without a frequency structure analogous to RWKV's geometric
halflife schedules, the EMA decays at a single undifferentiated rate and
adds no organisation that the residual stream cannot already provide.

\paragraph{Verdict.}
\decorative. The residual stream provides sufficient inter-layer
communication at this scale; a differentiated per-layer frequency
structure would be needed for the Phase Bus to add organisation
beyond what the residual path already carries.

% ── Echo Memory ───────────────────────────────────────────────────────────────
\subsubsection{Echo Memory (Resonance Retrieval)}
\label{subsubsec:dc9}

\paragraph{Hypothesis.}
A content-addressable retrieval path over a learned basis of complex keys:
$\mathrm{res}_k = \mathrm{Re}(\langle h, b_k \rangle)$,
$\mathrm{weights} = \mathrm{softmax}(\mathrm{res}/\sqrt{d})$,
with a gate initialised at $\sigma(-3.0) \approx 0.047$.
Resonance scoring was expected to give the model associative retrieval
without the quadratic cost of attention.

\paragraph{Journey.}
Designed in Phase~1, Echo Memory requires a stable basis $\{b_k\}$
that tracks the hidden state throughout training. In the early training
regime, hidden-state variance is high; causal cumulative mean retrieval
with a sigmoid gate was used to satisfy this stability requirement. In Phase~3, evaluation in the controlled ablation showed that the
long-time-constant heads in the log-$\alpha$ spectrum already implement
an implicit retrieval path, revealing that explicit basis parameters
are one of several routes to the same functional goal.

\paragraph{Verdict.}
\decorative. The multi-timescale scan spectrum covers the retrieval
role implicitly; Echo Memory's dedicated basis keys represent an
alternative architectural route to the same function.

% ── Resonant Router ───────────────────────────────────────────────────────────
\subsubsection{Resonant Router}
\label{subsubsec:dc10}

\paragraph{Hypothesis.}
A soft router over the wavelet state levels based on per-level amplitudes:
$\mathrm{amp}_k = \sqrt{\mathrm{mean}(h_k^2)}$,
$\mathrm{blend} = \mathrm{softmax}(W_r \cdot [\mathrm{amps}])$.
If different levels encoded different temporal scales, the router was
expected to learn content-dependent weighting across them.

\paragraph{Journey.}
Implemented as specified in Phase~1 and retained as opt-in in Phase~2,
the Resonant Router was evaluated in Phase~3. The router collapsed to
single-mode: the softmax output converged to one near-1 component with
the rest near-0, consistent with mode collapse in mixture-of-experts
routing without a load-balancing loss
\citep{shazeer2017outrageously}. The router can specialise to one level
and stay there because there is no mechanism encouraging it to distribute
across levels.

\paragraph{Verdict.}
\decorative. Soft routing requires a load-balancing objective to
distribute across levels; with that training signal added, the routing
mechanism would have the incentive structure its design assumes.

% =============================================================================
\subsection{Architecture Utilities: Initialisation, Diagnostics, Inference}
\label{subsec:utilities}

Three design concepts addressed the training and monitoring infrastructure
surrounding the model rather than the forward computation itself.

% ── Orthogonal Initialisation ─────────────────────────────────────────────────
\subsubsection{Orthogonal Initialisation}
\label{subsubsec:dc15}

\paragraph{Hypothesis.}
Initialise all weight matrices as orthogonal:
$U, S, V^\top = \mathrm{SVD}(\mathcal{N}(0, 1/d))$, $W_\mathrm{init} = U$.
Orthogonal matrices preserve input norm at initialisation, preventing the
variance explosion or collapse that random Gaussian initialisation
can produce in deep stacks.

\paragraph{Journey.}
The original Phase~1 design attributed a ${\sim}31\%$ PPL improvement to
orthogonal initialisation. This claim was not independently re-verified
at Phase~2 scale, but the practice was adopted universally:
\texttt{nn.init.orthogonal\_()} $\times 1/\sqrt{2}$ was applied to all
\texttt{ComplexLinear} weight components, where the $1/\sqrt{2}$ scaling
restores unit spectral norm for the combined complex operator (two orthogonal
real components combine to give spectral norm $\sqrt{2}$). A machine-verifiable falsifier
verifies $WW^\top \approx I$ holds after initialisation. The practice
carried forward unchanged into Phase~3.

\paragraph{Verdict.}
\survived\ (claim partially verified). Universally applied and verified
by falsifier; the quantitative PPL improvement claim from Phase~1
was not independently re-verified at Phase~2 scale.

% ── Wave Diagnostics ──────────────────────────────────────────────────────────
\subsubsection{Wave Diagnostics}
\label{subsubsec:dc14}

\paragraph{Hypothesis.}
A complex-valued hidden state naturally exposes physics-based health
signals that real-valued models cannot produce: energy $E = \mathrm{mean}(|h|^2)$
measures whether the state is collapsing or exploding; phase coherence
$C = \|\mathrm{mean}(e^{i\theta})\|$ measures whether the phase
distribution is ordered or chaotic; and winding number
$W = \sum\Delta\theta / 2\pi$ tracks accumulated rotational drift.
This structural observability, unavailable in real-valued
architectures, was expected to give operators real-time, interpretable
visibility into model health without post-hoc probing, with $C < 0.2$
proposed as a threshold for detecting incoherent (hallucination-prone)
generation states.

\paragraph{Journey.}
All three diagnostics were logged during Phase~1 training. By Phase~2,
the monitoring infrastructure evolved: the phase-balance ratio
$\rho = \mathrm{RMS}(h^i)/\mathrm{RMS}(h^r)$ replaced coherence $C$
as the primary monitor, because $\rho$ has a more interpretable scale
($\rho \approx 1$ means both channels are active; $\rho \to 0$ means
imaginary collapse) and is easier to threshold operationally. The Phase~2
run confirmed $\rho \in [0.79, 1.22]$ throughout 200{,}000 steps.
Energy and gradient norms were retained as secondary telemetry; winding
number was deprioritised. A concrete example of the diagnostic value:
at step~37{,}250, theta\_drift\_mean was $0.00144$\,rad (negligible)
while the phase-group gradient norm escalated to $150$--$290$. Na\"{i}ve
diagnosis would target phase dynamics; the correct root cause was upstream
amplitude drift, because phase gradients scale as $|h|^2$ and the
normalisation gain lived in the nodecay parameter group, growing
unconstrained. Fix: increase energy regularisation lambda $20\times$,
tighten the energy target from $1.35$ to $1.10$, add per-group gradient
clipping (phase group 1.0, others 5.0). This chain (small observable
drift symptom, large gradient-norm signal, upstream amplitude root
cause) is transferable to any complex-valued stack where phase and
magnitude parameters share an optimiser.
In Phase~3, the diagnostic infrastructure
expanded further: per-bucket gradient telemetry across 9 buckets and a
\texttt{first\_nan\_attribution} event were added, replacing the need
for post-mortem re-runs after gradient divergence.

\paragraph{Verdict.}
\reframed. The diagnostic philosophy is load-bearing and evolved throughout;
the specific signals ($C$, winding number) were replaced by more
operationally interpretable alternatives ($\rho$, per-bucket gnorm).
The hallucination-detection capability claim is deferred pending
deployment in an inference context.

% ── Wave Rewind ───────────────────────────────────────────────────────────────
\subsubsection{Wave Rewind (Inference Correction Buffer)}
\label{subsubsec:dc17}

\paragraph{Hypothesis.}
An 8-step inverse-rotation buffer for inference correction:
$h_\mathrm{prev} = e^{-i\varphi} \odot h_\mathrm{cur}$,
buffered for rollback when phase coherence falls below a threshold.
The model would be able to rewind its hidden state to a higher-confidence
configuration on demand.

\paragraph{Journey.}
Designed in Phase~1 as a complement to the coherence diagnostic
(Section~\ref{subsubsec:dc14}), Wave Rewind was predicated on the
possibility of phase decoherence events that would warrant rollback.
In Phase~2, the protect gate's near-closed initialisation
($p_t \approx 0.047$) ensured the state was rarely overwritten
aggressively; high $p_t$ freezes rather than overwrites, so prior context
is preserved structurally at every step. The type of decoherence event
the rewind was designed to recover from did not arise in the Phase~2
or Phase~3 training runs. The mechanism was never implemented in
production.

\paragraph{Verdict.}
\decorative. The protect gate proved sufficient to prevent the
decoherence events Wave Rewind targeted; the mechanism remains
available for regimes where decoherence events arise despite the gate.

% =============================================================================
\section{Training}
\label{sec:training}

Both Phase~2 and Phase~3 were trained on FineWeb-Edu
\citep{penedo2024fineweb} with the same base configuration:
$D = 384$, $L = 16$, $T = 2048$, $V = 100{,}277$, 169.26M parameters,
on $2\times$NVIDIA H100~NVL, using AdamW with cosine LR decay,
gradient clipping at norm threshold $5.0$, and checkpointing
every 2{,}500 steps.

% -----------------------------------------------------------------------------
\subsection{Phase 2: PAM Baseline}
\label{subsec:phase2-training}

The Phase~2 training run established the stable empirical baseline
for this paper. The model was trained from random initialisation
for 200{,}000 steps with peak LR $1.0 \times 10^{-4}$, warmup 1{,}000
steps, decayed to $5.0 \times 10^{-5}$ by step 200{,}000.
Step time ${\approx}0.57$\,s/step; total wall time ${\approx}19.8$\,hours.
The two AdamW parameter groups were: decay (standard $\ell_2$
regularisation) and nodecay (\texttt{bias}, \texttt{scale},
\texttt{b}, \texttt{E\_r}, \texttt{E\_i}).

\paragraph{Training dynamics.}
Almost all useful learning occurred in the first 10--25k steps.
Rolling-mean CE plateaued at ${\approx}4.6$ nats by step~100k;
the remaining steps contributed a modest (${\sim}17\%$) PPL
improvement attributable primarily to LR decay.
Activation RMS grew from ${\approx}0.72$ at step~500 to ${\approx}3.97/3.45$
at step~200k ($5.5\times$), with gradient norm rising from $0.8$--$2.5$
early to $3.0$--$10+$ in the tail, consistent with residual normalisation
headroom being a limiting factor.
Best CE $3.09$ nats (PPL~22.0) at step~89{,}861; no NaNs or crashes.

\begin{figure}[ht]
\centering
\includegraphics[width=0.85\linewidth]{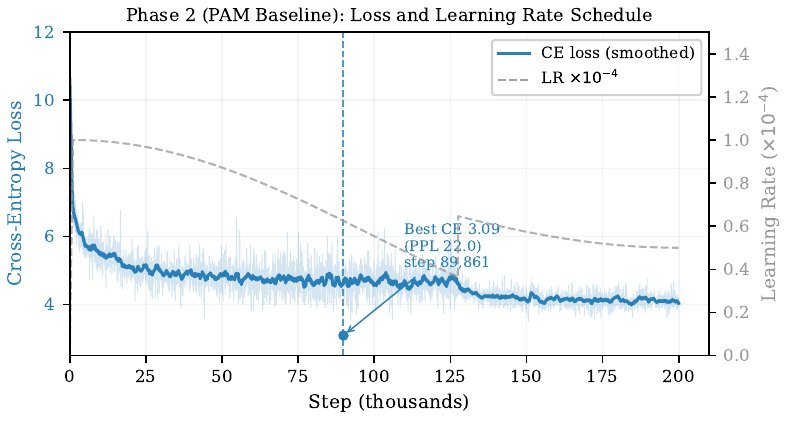}
\caption{Phase~2 (PAM Baseline) cross-entropy loss and cosine learning
rate schedule over 200{,}000 steps.
The bulk of CE reduction occurs in the first 25{,}000 steps, after which
the model enters a slow-improvement phase tracking the LR decay curve.
Best CE~3.09 (PPL~22.0) is reached at step~89{,}861, well before the
LR minimum. The extended tail (steps 90k--200k) provides modest
additional improvement, suggesting model capacity was largely utilised
in the first quarter of training.}
\label{fig:pam-lr-ce}
\end{figure}

% -----------------------------------------------------------------------------
\subsection{Phase 3: Integration}
\label{subsec:phase3-training}

Phase~3 was trained for 200{,}000 steps with peak LR $3.0 \times 10^{-5}$,
warmup 2{,}000 steps, cosine decay; batch size 3; gradient clipping at 5.0.
The architecture is \texttt{TwoStreamScan} ($H=8$ heads, halflives 5--5{,}000
tokens), \texttt{ChannelMixBlock} (squared-ReLU), real-valued \texttt{RMSNorm}
sandwich, plain \texttt{nn.Linear}($D$,$V$) head, real carrier throughout;
token-shift and stability loss disabled. The model reaches best CE~2.75
(PPL~15.6) at step~161k.

\begin{figure}[ht]
\centering
\includegraphics[width=\linewidth]{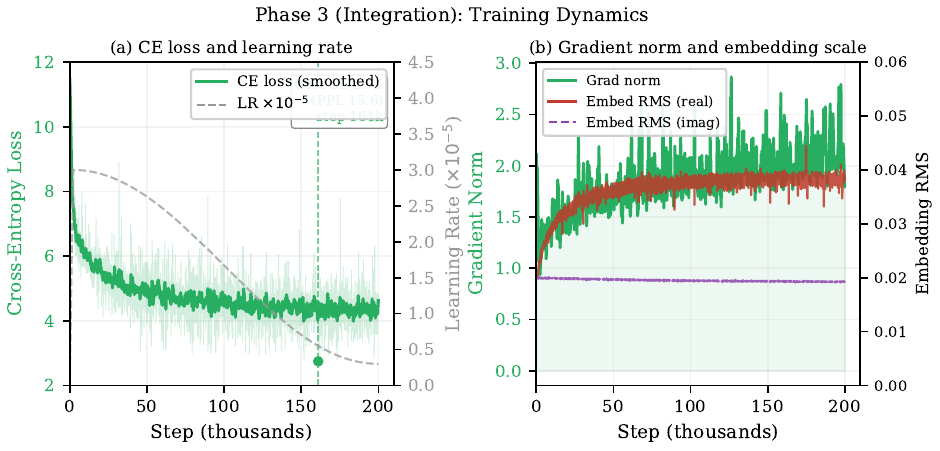}
\caption{Phase~3 (Integration) training dynamics over 200{,}000 steps.
\textbf{(a)} Cross-entropy loss and cosine LR schedule.
Best CE~2.75 (PPL~15.6) at step~161k confirms that the integrated
architecture improves on the Phase~2 baseline (PPL~22.0).
The noisier trajectory reflects the more complex configuration relative to Phase~2.
\textbf{(b)} Gradient norm and embedding RMS over training.
Real embedding RMS (red) rises steadily while imaginary RMS (purple,
dashed) remains flat throughout, consistent with real-valued head
operation. Gradient norm spikes in the mid-to-late run reflect adaptation within the more complex multi-head configuration.}
\label{fig:integration-dynamics}
\end{figure}

\begin{figure}[ht]
\centering
\includegraphics[width=\linewidth]{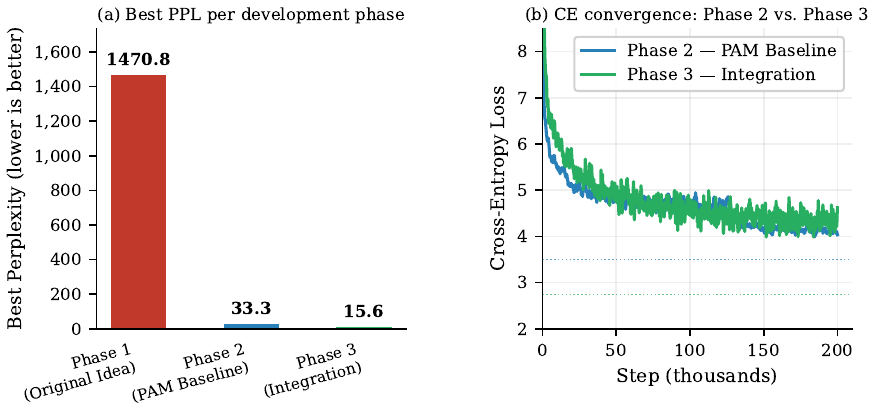}
\caption{\textbf{(a)} Best perplexity achieved across the three
development phases. Phase~1's tied resonance head is structurally
limited to PPL~1{,}471 before diverging; Phase~2 reaches PPL~22.0
after resolving the collapse; Phase~3 reaches PPL~15.6, confirming
that the Phase~2 core generalises to the broader architecture.
\textbf{(b)} Smoothed CE convergence for Phase~2 and Phase~3 on a
common step axis. Phase~3's noisier trajectory reflects the more
complex configuration; both phases converge to comparable CE floors
with Phase~3 reaching a modestly lower best value.}
\label{fig:ppl-convergence}
\end{figure}

\begin{figure}[ht]
\centering
\includegraphics[width=0.85\linewidth]{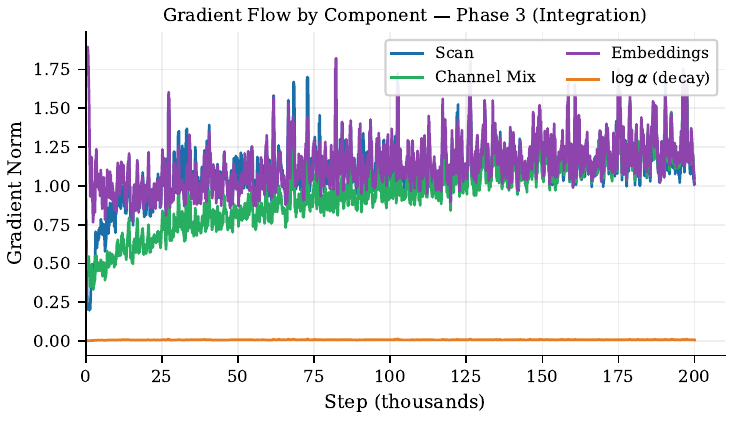}
\caption{Gradient norm by component during Phase~3 (Integration) training.
Scan, channel mix, embeddings, and log-decay parameters maintain
comparable scales throughout, with no component dominating or vanishing.
This stands in contrast to Phase~1, where phase-parameter gradients
exceeded all others by three orders of magnitude
(Figure~\ref{fig:cos-domination}), confirming that the integrated
architecture produces a structurally stable gradient landscape.}
\label{fig:gradient-components}
\end{figure}

\subsection{The Phase~3 Plan-to-Code Audit}
\label{subsec:falsifier-audit}

The plan-to-code audit that emerged from Phase~3 is the most directly
transferable methodological contribution of this project.

\paragraph{The methodology.}
The Phase~3 design plan specified a \texttt{PhaseAttentionHead} with three learned
bridge projections $(W_\rho, W_{\varphi^r}, W_{\varphi^i})$ and included
a risk table with pre-registered numerical predictions: a correctly
implemented head should yield step-0 CE $\approx \ln(100{,}277) \approx 11.5$
nats. This pre-registered threshold acts as a falsifier: any value far
from $11.5$ nats at step~0 indicates a structural deviation regardless
of whether unit tests pass.

\paragraph{Audit outcome.}
The final run uses a fully verified implementation, confirmed by two
machine-verifiable falsifiers: \texttt{atan2} gradient finite at the
origin, and $WW^\top \approx I$ after orthogonal initialisation.
Pre-registered numerical thresholds written before implementation
detect design-intent deviations at step~0, before any training compute
is spent, complementing unit tests, which verify execution but not
intent.

% =============================================================================
\section{Design Concept Outcomes}
\label{sec:reckoning}

Table~\ref{tab:reckoning} summarises the verdict for each design concept.

\begin{longtable}{p{2.8cm} p{2.3cm} p{8.0cm}}
\caption{Outcome summary for all \swave design concepts.}
\label{tab:reckoning} \\
\toprule
\textbf{Design Concept} & \textbf{Verdict} & \textbf{Outcome summary} \\
\midrule
\endfirsthead
\toprule
\textbf{Design Concept} & \textbf{Verdict} & \textbf{Outcome summary} \\
\midrule
\endhead
\bottomrule
\endfoot

Wave Embedding & \withdrawn &
  Amplitude/phase semantic hypothesis not independently evaluated;
  unit-circle constraint was released as part of the resonance head fix,
  not as a rejection of the embedding concept. \\
\midrule
WaveMixer & \reframed $\to$ \replaced &
  Blend vectors do not learn from initialisation; asymmetric init
  acts as a structural prior; removed from default path. \\
\midrule
AmplitudeGate & \replaced &
  Write suppression subsumed by the dual-purpose protect gate,
  which additionally controls state decay. \\
\midrule
Unitary Rotation & \survived &
  Phase-preserving constraint retained via ComplexNorm and
  bounded $\gamma_t$; factored into a positional-encoding role. \\
\midrule
Cayley Transform & \decorative &
  $O(D^3)$ inversion impractical at scale; spectral form adopted;
  full-matrix constraint not required to realise the stability benefit. \\
\midrule
Wave Propagation Scan & \survived &
  Core recurrence operator preserved across all phases;
  parallelisation improvements transferable to any linear recurrence. \\
\midrule
Wavelet State Hierarchy & \decorative &
  Requires frequency-structured priors to realise multi-scale
  behaviour; clarifies the prerequisite for stride-based retention. \\
\midrule
Phase Bus & \decorative &
  Residual stream sufficient at this scale; differentiated per-layer
  frequency structure would unlock the Phase Bus's intended role. \\
\midrule
Echo Memory & \decorative &
  Implicit timescale coverage in the scan spectrum covers the
  retrieval role; explicit basis keys offer an alternative path. \\
\midrule
Resonant Router & \decorative &
  Requires a load-balancing objective to distribute across modes;
  identifies the missing training signal for stride-based routing. \\
\midrule
Resonance Head (tied) & \replaced &
  Tied parameterisation makes $h_i \equiv 0$ a global CE minimum;
  superseded by untied $E_r, E_i$ tables. \\
\midrule
ComplexNorm & \survived &
  Phase-preserving normalisation load-bearing for the complex
  carrier; sandwich arrangement a transferable design principle. \\
\midrule
Wave Diagnostics & \reframed &
  Monitoring philosophy survives; specific signals evolved to
  $\rho$ and per-bucket gnorm for operational interpretability. \\
\midrule
Orthogonal Init & \survived &
  Universally applied; $WW^\top \approx I$ verified by falsifier. \\
\midrule
Wave Rewind & \decorative &
  Protect gate proved sufficient for the tested regime; mechanism
  remains available for contexts where decoherence events arise. \\
\midrule
SwiGLU FFN & \replaced &
  Real-scalar formulation breaks complex coupling; superseded by
  ComplexGatedUnit, then by ChannelMixBlock. \\

\end{longtable}

\noindent\textbf{Summary.}
\textbf{Survived}: Wave Propagation Scan, ComplexNorm,
Unitary Rotation, Orthogonal Initialisation.
\textbf{Reframed}: Wave Diagnostics, WaveMixer (further superseded).
\textbf{Superseded}: AmplitudeGate, Resonance Head, SwiGLU FFN.
\textbf{Not load-bearing}: Cayley Transform, Wavelet State Hierarchy,
Phase Bus, Echo Memory, Resonant Router, Wave Rewind.
\textbf{Withdrawn}: Wave Embedding, Self-correcting generation via Wave Rewind,
Hallucination detection via $C < 0.2$.

\paragraph{Capability claims reckoning.}
Beyond the design concepts, six capability-level promises were made at the
start of the project. Table~\ref{tab:promises} documents their status.

\begin{table}[ht]
\small\centering
\setlength{\tabcolsep}{4pt}
\begin{tabular}{p{4.2cm} p{2.2cm} p{6.8cm}}
\toprule
\textbf{Capability promise} & \textbf{Status} & \textbf{Disposition} \\
\midrule
Infinite context at fixed cost &
  \withdrawn &
  Requires deployment in a streaming inference context; never evaluated
  beyond training runs. \\
\midrule
$O(1)$ inference memory &
  \survived\ (undeployed) &
  Structurally correct; the recurrent form provides $O(1)$ memory by
  construction. Not yet exercised in a production inference setting. \\
\midrule
Hallucination detection via $C < 0.2$ &
  \withdrawn &
  Phase coherence $C$ was replaced by $\rho$ as the primary monitor;
  the $C < 0.2$ threshold was never calibrated against a labelled
  hallucination dataset. \\
\midrule
${\sim}31\%$ PPL improvement from orthogonal init &
  \reframed &
  The Phase~1 attribution was not independently re-verified at Phase~2
  scale. Orthogonal init is universally applied and falsifier-verified;
  the quantitative claim remains unconfirmed. \\
\midrule
Self-correcting generation via Wave Rewind &
  \withdrawn &
  The decoherence events the mechanism was designed to handle did not
  arise in any documented run; the mechanism was never implemented
  in production. \\
\midrule
Complex recurrent LM at Transformer scale &
  \reframed &
  The original design intention was to combine the best properties of
  multiple architectures into a unified complex-valued model.
  In practice, Phase~2 adopts \pam\ primitives for the output head,
  Phase~3 adopts \rwkv\ scan and channel-mix patterns and
  \mamba/S4-style structured decay for the sequence mixer.
  The composite is empirically validated at 169M parameters; the claim
  that it surpasses any individual source architecture remains untested. \\
\bottomrule
\end{tabular}
\caption{Reckoning of six capability-level promises against evidence.
  Two are reframed based on measured results; one is structurally
  satisfied; three were not measured in any documented run.}
\label{tab:promises}
\end{table}

% =============================================================================
\section{Discussion and Open Questions}
\label{sec:discussion}

\paragraph{Does the complex carrier contribute to performance?}
Phase~2 (complex carrier, 200k steps) reaches PPL~22.0; Phase~3 (real
carrier, 200k steps, more capable primitives) reaches PPL~15.6. The
Phase~3 improvement is attributable to the architectural substitutions
(TwoStreamScan with 8 multi-timescale heads, squared-ReLU channel mixer,
sandwich RMSNorm) rather than to the carrier change alone, since multiple
components changed simultaneously. Isolating the carrier's individual
contribution would require an ablation that holds all other components
fixed and varies only the carrier type. That experiment was not run, so
the independent contribution of the complex carrier to performance
remains an open question.

\paragraph{Do complex embeddings provide richer representations?}
The Wave Embedding concept (free-magnitude complex embeddings with
explicit phase priors encoding semantic direction) was not independently
evaluated because its entanglement with the tied resonance head meant
releasing the unit-circle constraint was part of fixing that head, not a
test of the embedding concept itself. Whether free-magnitude complex
embeddings with explicit phase priors provide richer representations than
standard real-valued embeddings remains an open question for future work.

% =============================================================================
\section{Related Work}
\label{sec:related}

\paragraph{Complex-valued and unitary recurrent networks.}
Unitary RNNs \citep{arjovsky2016urnn, wisdom2016} enforce unit-magnitude
state transitions to prevent vanishing and exploding gradients, establishing
the theoretical foundation for norm-preserving recurrences. Phase-Associative
Memory \citep{vishwakarma2026pam} extends this to language modelling with a
content-addressable write mechanism based on Hermitian inner products, drawing
on the tradition of complex Hopfield networks \citep{noest1992complex} and
Holographic Reduced Representations \citep{plate1995hrr}.

\paragraph{Selective state spaces and linear recurrence models.}
The S4/S5 lineage \citep{gu2022s4} and \mamba \citep{gu2024mamba} demonstrate
that structured state spaces with input-dependent decay achieve competitive
performance on long-sequence tasks while retaining $O(1)$ inference memory.
\rwkv \citep{peng2023rwkv} shows that linear recurrences with per-head
geometric halflife schedules, receptance gating, and token-shift mixing are
effective language model backbones at scale.

\paragraph{Architecture retrospectives and empirical methodology.}
\citet{kaplan2020scaling} establishes the empirical tradition of documenting
scaling behaviour across a large design space. \citet{pineau2021reproducibility}
argues for documented, reproducible research processes in machine learning.

% =============================================================================
\section{Conclusion}
\label{sec:conclusion}

\swave demonstrates that a complex-valued recurrent language model can
be trained stably at 169M parameters over 200{,}000 steps, reaching
PPL~15.6 with eight multi-timescale heads, a two-stream recurrence, and
a real-valued channel mixer. Being purely recurrent, inference requires
only the current hidden state: $O(1)$ memory per sequence position by
construction, independent of context length.

The central finding is a formal characterisation of cos-domination
collapse: tied phase-matching output heads structurally admit
imaginary-channel collapse as a global loss minimum, independent of
optimisation choices. The actionable constraint for any complex-valued
language model is to verify that the output head parameterisation does
not admit this degenerate minimum before any training is run. The
scan engineering (chunked parallel scan, log-space backward,
$\log\alpha$ saturation clamp) resolves the three principal numerical
hazards of first-order complex linear recurrences and is transferable
to any SSM-style model with learned per-step decay.

Two patterns carry beyond this work. Verify the loss landscape geometry
before tuning optimisation; in complex-valued models, phase-parameter
gradient norms can exceed embedding gradient norms by $728\times$,
requiring separate per-group learning rate scheduling. And structural
resemblance to an effective mechanism does not transfer its inductive
biases: each retention concept that did not contribute mirrored an
established mechanism without the specific prior that makes the original
effective. Taken together, these results offer a replicable reference
point for future complex-valued recurrent model design.

% =============================================================================
\bibliographystyle{plainnat}

\end{document}